*Article*

# Multiple Inputs and Mixed Data for Alzheimer's Disease Classification Based on 3D Vision Transformer

Juan A. Castro-Silva [1,2,*] , María N. Moreno-García [1] and Diego H. Peluffo-Ordóñez [3,4]

1   Data Mining (MIDA) Research Group, Universidad de Salamanca, 37007 Salamanca, Spain; mmg@usal.es
2   Faculty of Engineering, Universidad Surcolombiana, Neiva 410002, Colombia
3   Faculty of Engineering, Corporación Universitaria Autónoma de Nariño, Pasto 520001, Colombia; diego.peluffo@aunar.edu.co or peluffo.diego@um6p.ma
4   College of Computing, Mohammed VI Polytechnic University, Lot 660, Hay Moulay Rachid, Ben Guerir 43150, Morocco
*   Correspondence: juan.castro@usco.edu.co

**Abstract:** The current methods for diagnosing Alzheimer's Disease using Magnetic Resonance Imaging (MRI) have significant limitations. Many previous studies used 2D Transformers to analyze individual brain slices independently, potentially losing critical 3D contextual information. Region of interest-based models often focus on only a few brain regions despite Alzheimer's affecting multiple areas. Additionally, most classification models rely on a single test, whereas diagnosing Alzheimer's requires a multifaceted approach integrating diverse data sources for a more accurate assessment. This study introduces a novel methodology called the Multiple Inputs and Mixed Data 3D Vision Transformer (MIMD-3DVT). This method processes consecutive slices together to capture the feature dimensions and spatial information, fuses multiple 3D ROI imaging data inputs, and integrates mixed data from demographic factors, cognitive assessments, and brain imaging. The proposed methodology was experimentally evaluated using a combined dataset that included the Alzheimer's Disease Neuroimaging Initiative (ADNI), the Australian Imaging, Biomarker, and Lifestyle Flagship Study of Ageing (AIBL), and the Open Access Series of Imaging Studies (OASIS). Our MIMD-3DVT, utilizing single or multiple ROIs, achieved an accuracy of 97.14%, outperforming the state-of-the-art methods in distinguishing between Normal Cognition and Alzheimer's Disease.

## 1. Introduction

Alzheimer's Disease (AD) is a significant public health challenge and the leading cause of dementia in older adults. This progressive brain disorder leads to nerve cell death, substantial brain volume reduction, and the impairment of nearly all brain functions [1]. AD affects critical regions like the entorhinal cortex, fornix, hippocampus, and frontal, temporal, and parietal lobes, disrupting orientation, cognition, memory, intelligence, judgment, behavior, and language [2–4]. Consequently, early diagnosis, supportive care, and lifestyle interventions are essential for managing the disease and improving the quality of life.

Diagnosing Alzheimer's disease involves a multifaceted approach, including neurological exams, cognitive and functional assessments, brain imaging, and blood tests as no single test can definitively diagnose the disease. Cognitive tests like the Mini-Mental State Exam (MMSE) evaluate memory and problem-solving abilities. Imaging technologies, such as Magnetic Resonance Imaging (MRI) or Computed Tomography (CT), reveal brain conditions like tumors and strokes [5].

Deep-learning-based computer vision methods effectively detect brain structural changes via MRI using 2D or 3D models on entire images or specific regions of interest (ROIs). The AD classification models can be categorized by input type: 2D slices from 3D volumes, ROI-based models, 3D patches, and 3D subject-level models [6]. ROI-based models reduce memory use and overfitting risk but are limited by focusing on only a few brain regions, while AD affects multiple areas.

The transformer architecture [7], initially developed for sequence-to-sequence tasks like machine translation, is now widely used in NLP, computer vision, and speech processing. Vision Transformer (ViT) [8] models are increasingly applied in diagnosing Alzheimer's Disease due to their excellent performance. These models use self-attention to emphasize the connections between sequences, dividing images into uniform-sized patches embedded and processed by the transformer [9].

Further, 2D Transformers analyze slices independently, potentially missing contextual information and crucial 3D spatial data. In contrast, 3D Transformers process consecutive slices together, capturing the feature dimensions and spatial information across multiple scales through self-attention mechanisms [10].

The Alzheimer's Disease datasets include mixed data types such as age, MMSE scores, gender, and MRI scans. The deep learning models handling these data require preprocessing like scaling and normalization. Integrating multi-modal data is crucial for effective AD analysis as more than single modalities may be required. A method is necessary to effectively utilize information from diverse sources, harnessing the versatility of transformers and leveraging the rich contextual information provided by 3D imaging data [11].

The current methods for classifying AD face significant limitations, including the inadequate handling of mixed data types and insufficient integration of 3D contextual information. Many models rely on 2D slices from 3D volumes, potentially missing critical spatial details, or focus on a few brain regions, neglecting the widespread impact of AD. Additionally, comprehensively integrating diverse data sources remains a challenge. These issues highlight the need for a more comprehensive approach to AD classification.

To address these challenges, we propose a novel multiple-input transformer model. This model integrates categorical, numerical, and 3D ROI image data (including regions such as the entorhinal cortex, fornix, frontal lobe, hippocampus, parietal lobe, and temporal lobe) to enhance the performance and diagnostic accuracy. Our approach, "Multiple Inputs and Mixed Data for Alzheimer's Disease Classification based on 3D Vision Transformers", leverages the versatility of transformers and the rich contextual information from 3D imaging to improve the robustness and accuracy in diagnosing AD.

The proposed methodology consists of three main phases:

- Data Preparation (Phase 1): This phase involves splitting the dataset into training, validation, and test sets. Additionally, raw MRI volumes undergo skull-stripping and registration preprocessing to create refined datasets.
- Instance Selection (Phase 2): This phase involves selecting slice instances that contain the regions of interest (ROIs) and calculating the centroid position (x, y) based on the statistical mode, which identifies the most frequent values for accurate ROI localization.
- Model Validation (Phase 3): This phase involves batch data generation, training, and testing diverse classification models based on transformers. It also includes integrating multiple inputs by combining the categorical, numerical, and image data from different ROIs to enhance the model performance.

This study's novel contributions are the following:

1. Introducing a 3D Vision Transformer model that processes consecutive slices together, capturing the feature dimensions and spatial information through self-attention mechanisms. This model significantly improves Alzheimer's Disease diagnosis and classification by overcoming the limitations of 2D independent analysis.
2. Fusing multiple 3D ROI imaging data inputs from Alzheimer's Disease-affected areas helps to overcome the limitation of focusing on only one or a few brain regions. This approach enhances the model performance and robustness in Alzheimer's Disease diagnosis and classification.
3. Integrating mixed data from demographic factors, cognitive assessments, and brain imaging is essential because no single test can definitively diagnose Alzheimer's. This multifaceted approach improves the robustness and reliability of the diagnostic models.

The structure of the paper is outlined as follows: Section 2 presents some related works. The materials and methods used for preprocessing and building AD Transformer-based classification models are included in Section 3. Section 4 provides a detailed description of the experiments conducted in this work and the parameter settings used. The results of the experiments are discussed in Section 5. Section 6 addresses the limitations of the proposed methods. Finally, the concluding remarks of this work are summarized in Section 7.

## 2. Related Works

Vision and Swin Transformers have significantly advanced Alzheimer's Disease (AD) classification using MRI data. For instance, ref. [12] employs a Vision Transformer (ViT) pre-trained on ImageNet-21K for AD/CN classification. Similarly, ref. [13] combines a Vision Transformer with a 3D Siamese network to classify the MRI scans across various AD stages. Addressing AD-specific tasks, ref. [14] introduces a 3D Recurrent Visual Attention Model and an Attention Transformer. Additionally, ref. [15] proposes a Convolutional Voxel Vision Transformer (CVVT) for 3D MRI scans.

Examples of Swin Transformer applications include [16], which introduces the Resizer Swin Transformer (RST) for MRI brain image analysis. Further, ref. [17] develops ECSnet, a hybrid model combining CNN and Swin Transformer using 2.5D-subject data, and [18] presents a Conv-Swinformer, merging VGGNet-16's CNN module with a Swin Vision Transformer encoder. Lastly, ref. [19] proposes Trans-ResNet, integrating transformers and CNNs for advanced AD classification.

Earlier research often utilizes ROI-based models that focus on specific brain regions. For instance, [20] extracts hippocampus blocks from MRI scans, while [21] analyzes the medial temporal lobe using 30 coronal slices. Other approaches, such as [19,22,23], create ensemble classifiers by extracting patches from regions like the hippocampus, amygdala, and insulae. Further, [24] proposes a framework using ROIs and landmarks to streamline the attribute vectors, while [25] leverages Explainable AI and 3DCNN models for patient-specific ROI detection. Additionally, [26] introduces a method for extracting informative ROI content based on statistical modes.

Despite the advancements in Alzheimer's disease (AD) classification using deep learning, significant limitations persist. Many of the current methods rely on 2D Transformers that analyze individual MRI slices, missing crucial 3D context. ROI-based models focus on specific brain regions, overlooking the broader spectrum of AD impact. Additionally, most traditional classification models depend on a single type of test or data source, which fails to capture the multifaceted nature of the disease.

We propose a novel Multiple Inputs and Mixed Data 3D Vision Transformer to address these challenges. This approach integrates diverse data types: categorical, numerical, and 3D images from key AD-affected regions such as the entorhinal cortex, fornix, frontal lobe, hippocampus, parietal lobe, and temporal lobe. Our method aims to enhance the diagnostic performance, robustness, and accuracy, offering a more comprehensive tool for classifying Cognitively Normal (CN) and AD cases. By leveraging the versatility of transformers and the rich contextual information from 3D imaging, this model has the potential to improve the diagnostic accuracy and contribute to significantly better patient outcomes.

## 3. Materials and Methods

This section presents the dataset and the proposed methodology for building AD classification models. It covers data preparation, instance selection, ROI extraction, 3D ROI batch generation, and the transformer classification model used in this work.

## 3.1. Datasets

The datasets used in this study are sourced from (1) Alzheimer's Disease Neuroimaging Initiative (ADNI), a longitudinal multicenter study designed to develop clinical, imaging, genetic, and biochemical biomarkers for the early detection and tracking of AD [27], (2) Australian Imaging, Biomarker, and Lifestyle Flagship Study of Ageing (AIBL), a study to discover which biomarkers, cognitive characteristics, and health and lifestyle factors determine the subsequent development of symptomatic AD [28], and (3) Open Access Series of Imaging Studies (OASIS), a project aimed at making neuroimaging datasets of the brain freely available to the scientific community [29]. These are well-established and publicly available datasets widely used in Alzheimer's Disease research.

The dataset includes various data types, such as (a) 3D MRI Scans: Structural brain images focusing on specific regions of interest (ROIs) affected by Alzheimer's Disease. (b) Demographic Data: Information such as age, gender, and education level. (c) Cognitive Assessments: Scores from standardized cognitive tests used to assess cognitive function and decline.

Subjects in these datasets have been characterized using the Clinical Dementia Rating (CDR) scale [30], which is a measure that ranges from 0 to 3 and is used to determine the overall severity of dementia. A CDR of zero characterizes CN cases, while a CDR of one or greater represents AD cases.

Our study addresses prediction model performance across diverse demographic groups by integrating datasets from ADNI, AIBL, and OASIS, spanning various age groups (young and old). This results in a merged dataset of 420 instances, each contributing 70 volumes per class (3 datasets × 70 volumes × 2 classes). Demographic details are provided in Table 1.

**Table 1.** Summary of participant demographics and global clinical dementia rating (CDR) scores of all the study datasets.

| Dataset | Class | Subjects | Age | Gender F/M | Total Subjects |
|---|---|---|---|---|---|
| ADNI | CN | 70 | 78.63 ± 5.82 | 34/36 | 140 |
|  | AD | 70 | 78.63 ± 6.50 | 31/39 |  |
| AIBL | CN | 70 | 74.56 ± 5.81 | 37/33 | 140 |
|  | AD | 70 | 74.87 ± 7.57 | 43/27 |  |
| OASIS | CN | 70 | 69.89 ± 9.38 | 39/31 | 140 |
|  | AD | 70 | 76.36 ± 9.15 | 34/36 |  |
| MERGED | CN | 210 | 74.36 ± 8.01 | 110/100 | 420 |
|  | AD | 210 | 76.62 ± 7.93 | 108/102 |  |

## 3.2. Proposed Methodology

This methodology aims to create a multiple-input classification model based on transformers by combining image data from different regions of interest (ROIs) and mixed data to enhance performance, reliability, and robustness.

As illustrated in Figure 1, our methodology begins with data preparation, which includes splitting the dataset into training, validation, and test sets and preprocessing the images to remove non-informative content and align them properly. Next, the instance selection method identifies slices containing the defined ROIs and, based on the statistical mode, determines the centroid position (x, y). Finally, the proposed model is validated by generating batch data for training and testing diverse transformer-based classification models. This phase integrates multiple inputs by combining image data from different ROIs and mixed data.

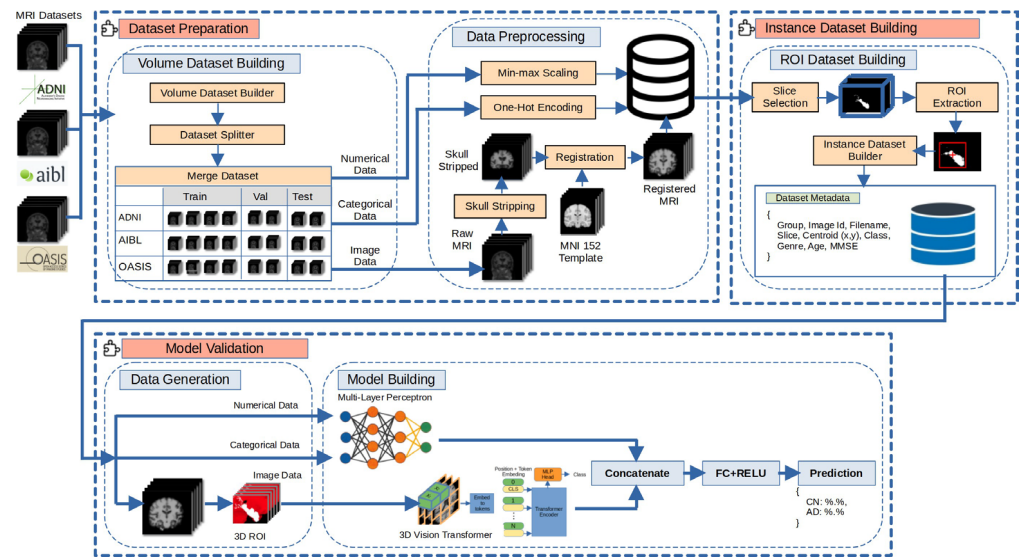

**Figure 1.** Proposed methodology.

3.2.1. Dataset Preparation

This phase selects one volume per subject and splits the dataset into training, validation, and test sets. Selected raw MRI volumes undergo skull-stripping and registration preprocessing.

Volume Dataset Building

Volumes per subject are sorted chronologically by visit date, and only the latest visit's volume is retained to ensure each subject contributes one volume. The datasets are then randomly split to ensure reproducible tests and prevent data leakage, with each MRI volume assigned exclusively to either the training, validation, or test set. Undersampling addresses class imbalance issues, ensuring an equal number of subjects per class (*k*), where *k* equals or is less than the number of samples from the minority class.

Data Preprocessing

Various preprocessing steps were applied to ensure data consistency and improve model performance in preparation for training. These steps included scaling numerical data, encoding categorical data, and specialized image data processing to maintain anatomical integrity and standardization across different datasets.

- Numerical Data: Min–max scaled to the range [0, 1] before training.
- Categorical Data: One-hot encoded, ensuring the resulting integer vectors were in the range [0, 1].
- Image Data:
    - Skull-Stripping: Raw volumes selected in the previous "Volume Dataset Building" step from ADNI, AIBL, or OASIS repositories underwent skull-stripping to isolate the brain from non-brain tissues such as skin, fat, muscle, neck, and eyeballs.
    - Registration: Skull-stripped dataset volumes were registered with the MNI152 T1 template MRI scan, ensuring uniformity in shape, position, and alignment. The resulting scans have dimensions of 182 × 218 × 182 and a resolution of 1 mm.
    - Scaling: Image data were scaled to the range [0, 1].

3.2.2. Instance Dataset Building

Instance selection techniques are crucial for optimizing prediction models by leveraging informative data, enhancing performance, minimizing computational costs, and reducing dataset size. This study adopts a novel instance selection method proposed in [26], which includes two critical approaches. Firstly, the Multi-Atlas ROI-based Instance

Selection integrates ROI annotations from multiple atlases to ensure informative slice retention and representativeness. Secondly, the ROI Content Extraction method utilizes the statistical mode to adjust the centroid (x, y) position, enabling precise extraction of the most informative content from ROIs for accurate 2D slice cropping.

3.2.3. Data Generation

Data generation begins by reading instance dataset metadata, including age, MMSE score, gender, volume filename, slice number, ROI centroid coordinates, and class labels. Data are loaded in batches to manage memory constraints. Before training, preprocessing steps include scaling numerical data to the range [0, 1] using min–max normalization, one-hot encoding categorical variables to produce integer vectors within [0, 1], and scaling image data to the range [0, 1]. Images are cropped to extract the desired 3D ROI.

3.2.4. Model Building

This step involves constructing transformer-based classification models using diverse ROIs and brain hemispheres. Additionally, it encompasses hyperparameter optimization and evaluating model performance, as detailed below:

Multiple-Input Network

- Multiple-Input Network: To build our Multiple Inputs and Mixed Data network, we will need two branches, as shown in Figure 2:
    - The first branch will be a simple Multi-layer Perceptron (MLP) designed to handle the categorical/numerical inputs.
    - The second branch will be a Swin Transformer to operate over the Magnetic Resonance image data.
    - These branches will then be concatenated together to form the final multiple-input model.

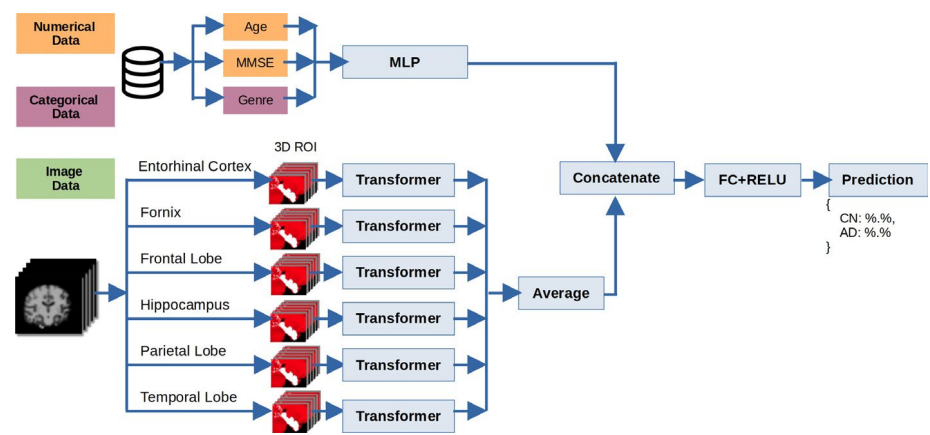

**Figure 2.** In the Multiple Inputs and Mixed Data model, categorical and numerical data serve as inputs to the MLP, while 3D ROI images are used as inputs to the 3D Vision Transformer.

- 3D Vision Transformer: This study uses ViViT: A Video Vision Transformer [31], a pure Transformer-based model, to classify AD MRI 3D ROIs. Figure 3 shows the Tubelet embedding method used to extract non-overlapping spatio-temporal "tubes" from the input volume and linearly project them to R$^d$. This method extends ViT's embedding to 3D and corresponds to a 3D convolution. For a tubelet of dimension $t \times h \times w$, tokens are extracted from the temporal, height, and width dimensions, intuitively fusing spatio-temporal information during tokenization [31].

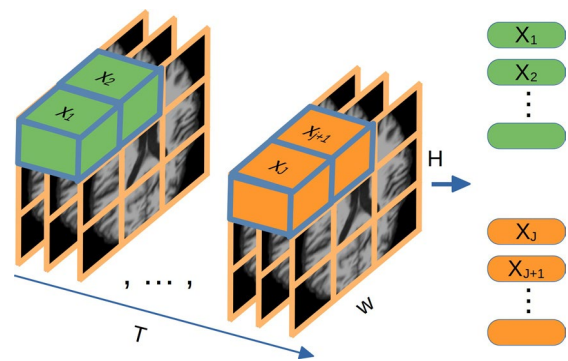

**Figure 3.** Tubelet embedding. Adapted from [31].

Hyperparameter Optimization

The effectiveness of machine learning algorithms hinges on identifying optimal hyperparameters. Hyperband, an advanced technique, optimizes random search by allocating predefined resources to randomly sampled configurations and implementing early stopping [32]. This study employs it to select the optimal hyperparameters for training the AD classification models.

Model Performance Evaluation

The Receiver Operating Characteristic (ROC) Curve illustrates the trade-off between the True Positive Rate (TPR) and False Positive Rate (FPR) at different classification thresholds. TPR (Sensitivity) indicates the proportion of actual positives correctly identified by the model, while FPR shows the proportion of actual negatives incorrectly classified as positives.

Due to limited data, the proposed model was validated using 7-fold cross-validation. This method ensures stable accuracy, prevents overfitting, and improves reliability by training on different data subsets. Cross-validation provides insights into model generalization and accurately estimates predictive performance on new data.

The overall performance of the binary classification model is quantified using the Area Under the Curve (AUC) metric, representing the area under the ROC curve, and the accuracy metric, represented as the average and standard deviation of the accuracies obtained from all seven folds.

## 4. Experimental Setup

### 4.1. Experiments

The proposed model was trained and evaluated using 3D ROI datasets from the entorhinal cortex, fornix, frontal lobe, hippocampus, parietal lobe, and temporal lobe. These datasets were created using the same MRI volumes to ensure reproducibility. The MRI volumes underwent preprocessing steps, including skull-stripping and registration to a common space.

#### 4.1.1. Mixed Data

This experiment evaluates the impact of mixed data on the proposed 3D Vision Transformer model's performance. The mixed data include genre (categorical), age, MMSE score (numerical), and 3D ROIs (image). The proposed model is trained on various ROIs affected by AD, such as the entorhinal cortex, fornix, hippocampus, frontal lobe, temporal lobe, and parietal lobe, as well as both brain hemispheres (left and right).

#### 4.1.2. Performance Comparison

This experiment compares the performance of the proposed Multiple Inputs and Mixed Data 3D Vision Transformer model with state-of-the-art methods. The related works analyzed in this experiment utilize diverse datasets (ADNI, AIBL, and OASIS), various model input types (2D and 3D), and different model architectures (Custom, Ensembles, Swin, and Vision Transformers).

*4.2. Hyperparameter Tuning*

Hyperband hyperparameter optimization is used to find optimal settings for training the Vision Transformer 3D model on the merged ADNI+AIBL+OASIS dataset. Table 2 summarizes key hyperparameter values for classifying AD cases from MRI scans.

**Table 2.** Hyperparameter values for 3D Vision Transformer merged dataset.

| Hyperparameter | Value | Description |
|---|---|---|
| Dataset | Merged | 70 Subjects each dataset (ADNI + AIBL + OASIS). |
| Slice Number | 25 | Number of selected slices from ROI. |
| Image Size | 32 × 32 | Image size in pixels. |
| Number of channels | 3 | Number of channels (3 = RGB, 1 = Gray scale). |
| Learning rate | LRS | LearningRateSchedule exponential decay. |
| Decay steps | 100,000 | Drops the learning rate by a factor. |
| Decay rate | 0.9 | The rate at which the learning rate is decayed. |
| Optimizer Name | Adam | The optimizer name. |
| Initial Learning Rate | $1e-04$ | The initial learning rate. |
| Dropout | 0.2 | The rate at which the learning rate is decayed. |
| Batch Size | 6 | The number of instances per batch. |
| Epochs | 250 | The number of epochs. |

*4.3. Data Analysis*

The one-way ANOVA test assessed significant differences between the means of three or more groups, and the T-test checked if two samples had identical averages. The significance level for this study was set at 0.05, or 5%.

In this work, the experiments have the below null and alternative hypotheses:

- $H_0$ (null hypothesis): $\mu_1 = \mu_2 = \mu_3 \cdots = \mu_k$ (It implies that the mean accuracy of all models is equal).
- $H_1$ (alternative hypothesis): It states that at least one model accuracy mean will differ from the rest.

*4.4. Implementation Details*

The subject dataset was split into training (70%), validation (15%), and test (15%) sets. The multiple-input model was implemented using Python software routines, with the random seed set for NumPy, TensorFlow, Random, and OS libraries to ensure reproducible results. Images were preprocessed using Python libraries NiBabel, TorchIO, PIL, and NumPy without saving them to disk. Skull-stripping and MRI registration with the MNI152 template were performed using FreeSurfer tools. The classification models were built using the Keras library. All models were trained at the Artificial Intelligence Laboratory, Universidad Surcolombiana (Neiva-Huila, Colombia), using ten HP OMEN Obelisk Desktop 875-102la workstations with Intel Core i9 9900K processors, 32 GB RAM, and 11 GB NVIDIA RTX 2080Ti GPUs.

**5. Results and Discussion**

The results correspond to the average and standard deviation of the 7-fold cross-validation.

*5.1. Experiment I (Mixed Data)*

The results presented in Table 3 demonstrate that using mixed data techniques (multimodal) significantly outperformed the only-image data methods (uni-modal) in terms of accuracy, with a *p*-value of 0.001, thus allowing us to reject the null hypothesis of equality.

The only-image data experiments utilize the 3D Vision Transformer, as shown in the second branch of Figure 2. In contrast, the mixed data experiments employ our proposed Multiple Inputs and Mixed Data 3D Transformer. Both sets of experiments are trained using only a single ROI.

**Table 3.** Summary of accuracy for 3D Vision Transformers from different ROIs and hemispheres using only-image and mixed data. The best performance models per ROI are highlighted.

|  | Only-Image Data | | Mixed Data | |
| --- | --- | --- | --- | --- |
| **ROI** | **Left** | **Right** | **Left** | **Right** |
| Entorhinal Cortex | 91.90 ± 2.02 | 90.00 ± 2.72 | 97.86 ± 0.81 | 95.95 ± 4.99 |
| Fornix | 89.76 ± 2.24 | 93.81 ± 1.59 | 97.86 ± 0.81 | 98.33 ± 0.00 |
| Frontal Lobe | 86.90 ± 2.44 | 89.52 ± 0.81 | 98.10 ± 0.63 | 98.10 ± 1.15 |
| Hippocampus | 88.57 ± 1.78 | 89.76 ± 2.62 | 97.86 ± 0.81 | 97.62 ± 1.89 |
| Parietal Lobe | 92.62 ± 2.33 | 89.76 ± 1.78 | 98.10 ± 0.63 | 98.33 ± 0.96 |
| Temporal Lobe | 89.29 ± 3.02 | 90.24 ± 2.24 | 98.33 ± 0.00 | 95.95 ± 5.60 |
| Mean | 89.84 ± 2.94 | 90.52 ± 2.46 | 98.02 ± 0.66 | 97.38 ± 3.19 |

Our proposed mixed data approach substantially improved the mean accuracy across both hemispheres. Specifically, it achieved increases of 8.18% for the left hemisphere and 6.86% for the right hemisphere compared to the only-image data method, with *p*-values of 0.001.

The mean accuracy of the ROIs in the only-image models exhibits significant differences. The lowest accuracy is found in the frontal lobe (left hemisphere), while the highest is in the fornix (right hemisphere), showing a difference of 6.91% with a *p*-value of 0.001.

Our methodology reveals a significant difference between the left and right hemispheres. Similarly, in [22], notable disparities are observed in the accuracy of the left and right hemispheres across three ROIs: the amygdala, hippocampus, and insula.

This approach significantly enhances the model robustness and reliability by integrating diverse data sources. Combining the data from demographic factors, cognitive assessments, and brain imaging ensures a multifaceted approach. These findings suggest that the proposed mixed data method could improve Alzheimer's Disease and Normal Cognition diagnosis.

*5.2. Experiment II (Performance Comparison)*

Given the novelty of 3D Vision Transformers, few existing methods utilize merged datasets, incorporate multiple inputs, and integrate mixed data for evaluating AD classification models. However, to benchmark the performance against the state-of-the-art approaches, Table 4 presents the classification results of various transformer-based methods from the recent literature across different datasets. Our best results from Table 3 are included for immediate comparison.

The experimental results indicate that the proposed Multiple Inputs and Mixed Data 3D Vision Transformer method, achieving 97.14% accuracy using multiple ROIs, and the single ROI classifiers—Entorhinal Cortex-Left (97.86%), Fornix-Right (98.33%), Frontal Lobe-Left (98.10%), Hippocampus-Left (97.86%), Parietal Lobe-Right (98.33%), and Temporal Lobe-Left (98.33%)—slightly outperform the best related work [12], which achieved 96.80% accuracy using a Vision Transformer method. These results demonstrate the effectiveness of our approach in enhancing the classification performance for Alzheimer's Disease diagnosis.

The superior performance can be attributed to implementing a 3D Vision Transformer model that processes consecutive slices together, capturing the feature dimensions and spatial information through self-attention mechanisms. Additionally, fusing multiple 3D ROI imaging data inputs from Alzheimer's Disease-affected brain areas and integrating mixed data from demographic factors, cognitive assessments, and brain imaging contribute to the robustness and accuracy of the model.

**Table 4.** Performance comparison of the proposed method with other related works for the classification tasks (AD vs. NC). Our methodology's performance using multiple ROIs is highlighted.

| Study | Dataset | Model | Accuracy |
|---|---|---|---|
| [19] | ADNI | Trans-ResNet | 93.85 |
|  | AIBL |  | 93.17 |
| [12] | ADNI | Vision Transformer | 96.80 |
| [18] | ADNI, OASIS | 2D CNN+Transformer | 93.56 |
| [14] | OASIS | Vision Transformer | 91.18 |
| [17] | ADNI, AIBL | CNN+Swin Transformer | 93.90 |
| [16] | ADNI+AIBL | Swin Transformer | 94.05 |
| [13] | ADNI+OASIS | Vision Transformer | 89.02 |
| Our Methodology Multiple ROIs | Merged (ADNI + AIBL + OASIS) | Multiple Inputs and Mixed Data 3D Vision Transformer | 97.14 |
| Single ROI |  | Entorhinal Cortex-Left | 97.86 |
|  |  | Fornix-Right | 98.33 |
|  |  | Frontal Lobe-Left | 98.10 |
|  |  | Hippocampus-Left | 97.86 |
|  |  | Parietal Lobe-Right | 98.33 |
|  |  | Temporal Lobe-Left | 98.33 |

The ROC curves and AUC values in Figure 4a demonstrate the effectiveness of the proposed Multiple Inputs and Mixed Data 3D Vision Transformer. The high average AUC value across 7-folds (0.984 ± 0.011) reflects the outstanding model performance and a solid capability to differentiate between CN and AD classes. The confusion matrix shown in Figure 4b indicates that the model performs well, accurately predicting most cases with minimal errors, reflecting its effectiveness in distinguishing between the CN and AD classes.

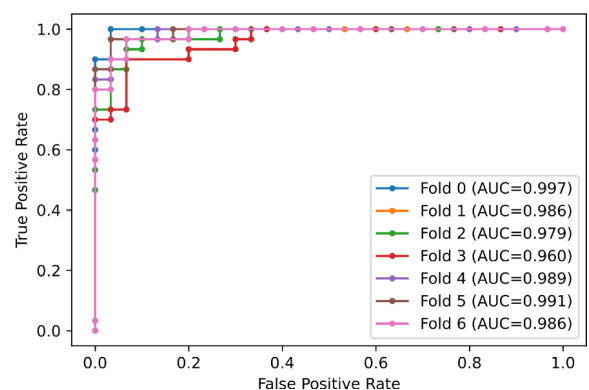
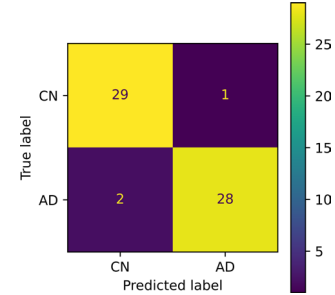

(**a**) ROC Curve (7 Folds)　　　(**b**) Confusion Matrix (Fold 0)

**Figure 4.** Performance evaluation: (**a**) ROC curve and (**b**) confusion matrix.

## 6. Limitations

The proposed method, which primarily relies on three data modalities available in public repositories, has the potential to partially replicate the extensive range of diagnostic tools used by clinicians in practice. In typical clinical settings, doctors employ a wide variety of diagnostic tools, including neurological exams, comprehensive cognitive and functional assessments, various brain imaging techniques, and blood tests, all of which contribute to a more holistic view of the patient's condition.

Another significant limitation of our study is the relatively small dataset size, which is particularly challenging for the 3D Vision Transformer models. These models demand a large amount of data to avoid overfitting and effectively learn and capture the complex 3D features in imaging data.

## 7. Conclusions and Future Work

This study introduces a novel methodology called the Multiple Inputs and Mixed Data 3D Vision Transformer (MIMD-3DVT). This method processes consecutive slices together to capture the feature dimensions and spatial information, fuses multiple 3D ROI imaging data inputs, and integrates mixed data from demographic factors, cognitive assessments, and brain imaging.

We propose a multi-source domain adaptation method to address the domain shift and enhance the model robustness. This approach consolidates three diverse population datasets (ADNI, AIBL, and OASIS) into a unified domain, creating a multicenter dataset with registered MRI skull-stripped images. The subjects are characterized using the Clinical Dementia Rating (CDR) scale for integrated data comparison.

Experimentally, the proposed methodology's impact on the performance of transformer-based classification models is explored. Our MIMD-3DVT, utilizing single or multiple ROIs, slightly outperforms the state-of-the-art methods in classifying Alzheimer's Disease.

For future work, integrating multiple data modalities, such as PET images, gray and white matter segmented images, brain volumetric data, and other types of exams, could significantly enhance the performance and reliability of the classification models. Additionally, increasing the dataset size is crucial for improving the performance and accuracy, enhancing generalization to new data, and ensuring a more diverse and representative sample. This approach will capture a broader range of patterns and features associated with Alzheimer's Disease, thereby improving the robustness of the diagnostic models.

**Author Contributions:** Conceptualization, J.A.C.-S.; Methodology, J.A.C.-S., M.N.M.-G. and D.H.P.-O.; Software, J.A.C.-S.; Validation, J.A.C.-S.; Formal analysis, J.A.C.-S. and D.H.P.-O.; Investigation, J.A.C.-S. and M.N.M.-G.; Visualization, J.A.C.-S. and D.H.P.-O.; Writing—review & editing, M.N.M.-G.; Supervision, M.N.M.-G.; Writing J.A.C.-S. All authors have read and agreed to the published version of the manuscript.

**Funding:** This research received no external funding.

**Data Availability Statement:** The datasets used in this manuscript are publicly available in the OASIS, AIBL, and ADNI repositories and can be accessed via the following links: https://www.oasis-brains.org/ and https://adni.loni.usc.edu/.

**Acknowledgments:** The authors are grateful for the support provided by the SDAS Research Group (https://sdas-group.com/). They would like to express their deep gratitude to Lorena Guachi for her invaluable insights and support. The data used in this article were obtained from the Alzheimer's Disease Neuroimaging Initiative (ADNI) database adni.loni.usc.edu (accessed on 11 May 2022). The investigators within the ADNI contributed to the design and implementation of ADNI and/or provided data but did not participate in the analysis or writing of this report. A complete listing of ADNI investigators can be found at http://adni.loni.usc.edu/wp-content/uploads/how_to_apply/ADNI_Acknowledgment_List.pdf (accessed on 11 May 2022). Data used in the preparation of this article were also partly obtained from the Australian Imaging, Biomarker, and Lifestyle Flagship Study of Ageing (AIBL), funded by the Commonwealth Scientific and Industrial Research Organization (CSIRO), and made available in the ADNI database www.loni.usc.edu/ADNI (accessed on 11 May 2022). The AIBL researchers contributed data but did not participate in the analysis or writing of this report. AIBL researchers are listed at https://aibl.csiro.au/ (accessed on 11 May 2022).

**Conflicts of Interest:** The authors declare no conflict of interest.